\documentclass[11pt]{article}

\usepackage[margin=1in]{geometry}
\usepackage[utf8]{inputenc}
\usepackage[T1]{fontenc}
\usepackage{hyperref}
\usepackage{url}
\usepackage{booktabs}
\usepackage{amsfonts}
\usepackage{nicefrac}
\usepackage{xcolor}
\usepackage{algorithm}
\usepackage{algorithmic}
\usepackage{graphicx}
\usepackage{amsmath}
\usepackage{amssymb}
\usepackage{multirow}
\usepackage{subcaption}
\usepackage{natbib}

\title{GATS: Graph-Augmented Tree Search with Layered World Models for Efficient Agent Planning}

\author{
  Maureese Williams\\
  \texttt{maureesewilliams@gmail.com}\\
	Dymitr Nowicki\\
	Institute for Cybernetics of NAS of Ukraine\\
	\texttt{nowicki@nnteam.org.ua}
}

\begin{document}

\maketitle

\begin{abstract}
Large Language Model (LLM) agents have shown promise in multi-step planning tasks, but existing approaches like LATS (Language Agent Tree Search) and ReAct rely heavily on LLM inference during planning, leading to high computational costs and stochastic behavior. We present \textbf{GATS} (Graph-Augmented Tree Search), a planning framework that combines systematic UCB1-based tree search with a layered world model to eliminate LLM calls during inference while achieving superior planning performance. Our three-layer world model integrates: (L1) exact symbolic action matching, (L2) statistics learned from execution logs, and (L3) LLM-based prediction for unknown actions. On synthetic planning tasks with branching paths and dead-ends, GATS achieves \textbf{100\% success rate} compared to 92\% for LATS and 64\% for ReAct. On a comprehensive stress test spanning 12 challenging scenarios---including coding workflows, web navigation, and long-horizon tasks---GATS maintains \textbf{100\% success} while LATS drops to 88.9\% and ReAct to 23.9\%. GATS requires \textbf{zero LLM calls per task} during planning (vs. 37 per task for LATS) and produces deterministic plans with zero variance across runs. Our results demonstrate that systematic search with learned world models can substantially outperform LLM-guided exploration for agent planning.
\end{abstract}

\section{Introduction}

The emergence of Large Language Models (LLMs) as reasoning engines has sparked significant interest in LLM-based agents capable of multi-step planning and tool use \citep{yao2022react, shinn2023reflexion, yao2023tree}. These agents must navigate complex decision spaces, often with partial information, dead-ends, and long-horizon dependencies. While LLMs provide powerful semantic understanding, directly using them for planning faces two key challenges: (1) \textit{computational cost}---each planning step requires expensive LLM inference, and (2) \textit{stochasticity}---LLM sampling introduces variance that makes plans non-reproducible.

Recent work has explored tree search methods for LLM agents. LATS (Language Agent Tree Search) \citep{zhou2023lats} combines Monte Carlo Tree Search with LLM-based action proposal and value estimation. While effective, LATS requires LLM calls at every search node, making it computationally expensive. Tree of Thoughts (ToT) \citep{yao2023tree} similarly relies on LLM evaluation for branch selection.

We propose \textbf{GATS} (Graph-Augmented Tree Search), a planning framework that decouples the world model from the LLM, enabling systematic search without inference-time LLM calls. Our key insight is that action effects in many domains can be captured by a \textit{layered world model}:

\begin{itemize}
    \item \textbf{L1 (Symbolic)}: Exact precondition-effect matching for known actions
    \item \textbf{L2 (Learned)}: Statistical predictions from execution logs
    \item \textbf{L3 (Generative)}: LLM-based prediction for novel situations
\end{itemize}

During planning, GATS uses UCB1-based tree search \citep{kocsis2006ucb} with the world model for state prediction, falling back through layers as needed. The LLM (L3) is only invoked for genuinely unknown actions, which are then cached for future use.

Our contributions are:
\begin{enumerate}
    \item A \textbf{layered world model architecture} that combines symbolic, learned, and generative components for efficient state prediction.
    \item A \textbf{systematic UCB1-based search} algorithm that outperforms random LLM-guided exploration.
    \item \textbf{Comprehensive evaluation} on 100 synthetic planning tasks and a 12-category stress test (120 tasks), showing GATS achieves 100\% success rate with zero LLM calls, compared to 92\% for LATS (main) and 88.9\% (stress test) with $\sim$37 LLM calls per task.
    \item \textbf{Ablation studies} demonstrating budget scaling and world model layer contributions.
\end{enumerate}

\section{Related Work}

\paragraph{LLM-Based Agents}
ReAct \citep{yao2022react} interleaves reasoning and acting, using the LLM to select actions based on observations. Reflexion \citep{shinn2023reflexion} adds self-reflection for learning from failures. Voyager \citep{wang2023voyager} demonstrates open-ended exploration in Minecraft using LLM-driven skill acquisition. DEPS \citep{wang2023deps} proposes describe-explain-plan-select for complex task decomposition. These methods rely on the LLM for every decision, limiting scalability.

\paragraph{Tree Search for LLMs}
Tree of Thoughts (ToT) \citep{yao2023tree} explores multiple reasoning paths using BFS/DFS with LLM evaluation. LATS \citep{zhou2023lats} applies MCTS with LLM-based action proposal and value estimation. RAP \citep{hao2023reasoning} uses world models for MCTS but still requires LLM calls for state transitions. Recent work on LLM reasoning \citep{lightman2023verify} shows the value of process-based verification. Our work differs by using a pre-computed layered world model that eliminates inference-time LLM dependence.

\paragraph{World Models}
World models have been successful in reinforcement learning \citep{ha2018world, hafner2019dream}. Recent work has explored LLM-based world models \citep{hao2023reasoning}, but these still require LLM inference per prediction. Our layered approach uses the LLM only for bootstrapping, then relies on symbolic and learned components.

\paragraph{Classical Planning}
GATS draws inspiration from classical AI planning \citep{ghallab2004automated}, particularly STRIPS-style action representations and heuristic search. We combine these with modern LLM capabilities for handling novel actions.

\section{Method}

\subsection{Problem Formulation}

We consider planning problems defined by:
\begin{itemize}
    \item State space $\mathcal{S}$: Sets of propositions (facts)
    \item Action space $\mathcal{A}$: Actions with preconditions and effects
    \item Initial state $s_0 \in \mathcal{S}$
    \item Goal condition $G \subseteq s$ for goal states
\end{itemize}

An action $a \in \mathcal{A}$ is \textit{applicable} in state $s$ if $\text{prec}(a) \subseteq s$. Applying $a$ yields state $s' = (s \cup \text{add}(a)) \setminus \text{del}(a)$. The objective is to find an action sequence $\pi = (a_1, \ldots, a_n)$ such that executing $\pi$ from $s_0$ reaches a goal state.

\subsection{Layered World Model}

The world model $\mathcal{W}$ predicts the next state and confidence given current state and action: $\mathcal{W}(s, a) \rightarrow (s', p)$. Our three-layer architecture queries layers in order until a confident prediction is obtained:

\paragraph{L1: Symbolic Matching}
For actions with known STRIPS-style specifications:
\begin{equation}
    \mathcal{W}_1(s, a) = \begin{cases}
        (s', 1.0) & \text{if } \text{prec}(a) \subseteq s \\
        (s, 0.0) & \text{otherwise}
    \end{cases}
\end{equation}
where $s' = (s \cup \text{add}(a)) \setminus \text{del}(a)$.

\paragraph{L2: Learned Statistics}
For actions observed during execution but without formal specifications:
\begin{equation}
    \mathcal{W}_2(s, a) = \left(\text{argmax}_{e} \text{count}(a \rightarrow e), \min\left(\frac{\text{count}(a)}{10}, 1.0\right)\right)
\end{equation}
This layer maintains transition statistics from execution logs, returning the most frequent effect with confidence $c = \min(\text{count}(a)/10, 1.0)$, saturating at 1.0 after 10 observations.

\paragraph{L3: LLM Prediction}
For novel actions, we query the LLM:
\begin{equation}
    \mathcal{W}_3(s, a) = (\text{LLM}(s, a), 0.5)
\end{equation}
Predictions are cached to avoid repeated LLM calls. In practice, L3 is rarely invoked when L1/L2 have sufficient coverage.

%

\paragraph{Layer Selection and Cost Model}
The layers are ordered by increasing generality and computational cost. Let
$c_1$, $c_2$, and $c_3$ denote the cost of querying L1, L2, and L3, respectively,
with $c_1 \ll c_2 \ll c_3$. In practice, $c_1$ is a symbolic lookup, $c_2$ is a
statistics lookup, and $c_3$ is an LLM inference call. For a transition query
$(s,a)$, the expected cost of the layered world model is
\begin{equation}
	\mathbb{E}[C_{\mathcal{W}}]
	=
	P(L_1)c_1 + P(L_2)c_2 + P(L_3)c_3 .
\end{equation}
Since L3 corresponds to an LLM call, the main efficiency objective is to maximize
coverage by L1 and L2 and to invoke L3 only for genuinely novel or ambiguous
transitions.

We write the layer-selection rule as
\begin{equation}
	\mathcal{W}(s,a) =
	\begin{cases}
		\mathcal{W}_1(s,a), & \text{if } p_1(s,a) \geq \tau_1,\\
		\mathcal{W}_2(s,a), & \text{if } p_2(s,a) \geq \tau_2,\\
		\mathcal{W}_3(s,a), & \text{otherwise}.
	\end{cases}
\end{equation}
Here $p_i(s,a)$ is the confidence assigned by layer $L_i$, and $\tau_i$ is the
minimum confidence required to accept that layer's prediction. In our
experiments, L1 predictions have confidence $p_1=1$ when the action is
applicable and $p_1=0$ otherwise. L2 confidence is estimated from observed
transition counts:
\begin{equation}
	p_2(a) = \min\left(\frac{n(a)}{n_0}, 1\right),
\end{equation}
where $n(a)$ is the number of observed executions of action $a$ and $n_0$ is a
saturation threshold. This count-based estimator reflects the intuition that
repeated consistent observations should increase trust in the learned transition
model.

\paragraph{LLM-Call Amortization}
Let $B$ be the search budget per planning step and $D$ the planning horizon.
A planner that calls the LLM during every node evaluation has cost approximately
\begin{equation}
	C_{\mathrm{LLM\text{-}search}} = O(BD \cdot c_3).
\end{equation}
By contrast, GATS caches L3 predictions:
\begin{equation}
	\mathrm{Cache}[k(s),a] = (s',p),
\end{equation}
where $k(s)$ is a canonical representation of state $s$. Therefore, the number
of LLM calls made by GATS is bounded by the number of distinct transition
queries for which L1 and L2 fail:
\begin{equation}
	N_{\mathrm{LLM}}^{\mathrm{GATS}}
	=
	\left|
	\left\{
		(k(s),a):
		L_1 \text{ fails},\,
		L_2 \text{ fails},\,
		(k(s),a) \notin \mathrm{Cache}
		\right\}
	\right|.
\end{equation}
This changes the role of the LLM from repeated planner-in-the-loop inference to
one-time transition-model bootstrapping.

\subsection{State-Transition Graph Memory}

The graph component of GATS is represented explicitly as a persistent state-transition graph
\begin{equation}
	\mathcal{G} = (\mathcal{V}, \mathcal{E}).
\end{equation}
Each node $v \in \mathcal{V}$ corresponds to a canonical state key $k(s)$, and
each directed edge corresponds to a predicted transition:
\begin{equation}
	e = (k(s), a, k(s')) \in \mathcal{E},
	\qquad
	(s',p) = \mathcal{W}(s,a).
\end{equation}
The edge stores the action $a$, the predicted successor state $s'$, the
world-model confidence $p$, the layer used to generate the prediction, and
search statistics such as visit count and accumulated value.

This graph representation differs from ordinary tree search in one important
way: if two different action sequences reach the same state, they are merged:
\begin{equation}
	k(s_i) = k(s_j) \quad \Rightarrow \quad v_i = v_j .
\end{equation}
Thus, transition predictions, value estimates, and visit counts are reused
across paths. This is especially useful in tool-use domains where different
sequences of actions can lead to the same intermediate configuration.

The graph is maintained across planning steps. After executing an action and
moving to a new state, GATS does not discard previous search statistics; instead,
it continues planning from the corresponding graph node. This amortizes both
world-model prediction and value estimation across the full episode.

The graph can also support dead-end detection. A node is marked as a dead end
when no path from that node reaches a goal node under the currently known
transition graph. Actions leading into known dead-end regions are assigned low
value during subsequent search.

Unlike ordinary tree search, which creates a separate node for every action
sequence, the state-transition graph stores each reachable state only once. If
two different partial plans lead to the same canonical state, their nodes are
merged and the planner reuses the same outgoing edges, transition predictions,
visit counts, and value estimates. In tree search, these two histories would be
expanded independently, duplicating both world-model queries and downstream
value estimation. In GATS, the canonical state key $k(s)$ acts as a
transposition table: all paths reaching $s$ share the same graph node and
therefore share accumulated search evidence.

The graph is also persistent across planning steps. After the planner selects
and executes an action, it does not discard the search structure built at the
previous step. Instead, planning resumes from the graph node corresponding to
the new state, while the rest of $\mathcal{G}$ remains available for future
reuse. This persistence is important in multi-step tool-use problems, where
similar states can be encountered repeatedly after different action sequences or
after recovering from suboptimal choices. As a result, GATS amortizes not only
LLM-based transition prediction but also symbolic transition lookup, BFS-based
value estimation, and UCB statistics over the full episode rather than over a
single local search call.

\subsection{UCB Selection over Graph Edges}

At a given state $s$, each applicable action $a \in A_{\mathrm{app}}(s)$ is
treated as an arm in a finite-armed bandit problem. Pulling an arm corresponds
to expanding or reusing the graph edge $(k(s),a,k(s'))$, predicting the successor
state using the layered world model, and evaluating the resulting state.

Let $N(s,a)$ be the number of times action $a$ has been evaluated from state $s$,
and let $Q(s,a)$ be its empirical mean value:
\begin{equation}
	Q(s,a)
	=
	\frac{1}{N(s,a)}
	\sum_{i=1}^{N(s,a)} v_i .
\end{equation}
GATS selects actions using a UCB rule:
\begin{equation}
	a_t =
	\arg\max_{a \in A_{\mathrm{app}}(s)}
	\left[
	Q(s,a)
	+
	c\sqrt{\frac{2\ln N(s)}{N(s,a)}}
	\right],
\end{equation}
where
\begin{equation}
	N(s) = \sum_{a \in A_{\mathrm{app}}(s)} N(s,a).
\end{equation}
For unvisited actions, we define
\begin{equation}
	N(s,a)=0
	\quad \Rightarrow \quad
	\mathrm{UCB}(s,a)=+\infty .
\end{equation}
This guarantees that every applicable action is evaluated at least once before
the algorithm commits budget to apparently high-value actions.

The value backed up through an edge can incorporate world-model confidence:
\begin{equation}
	v = p \cdot \mathrm{StateValue}(s') - \lambda(1-p),
\end{equation}
where $p$ is the confidence returned by the selected world-model layer and
$\lambda \geq 0$ penalizes uncertain predictions. In the fully symbolic setting,
$p=1$, and the update reduces to the deterministic value backup used in
classical planning.

\begin{algorithm}[t]
	\caption{GATS Search with Explicit Graph Memory}
	\label{alg:gats_graph_search}
	\begin{algorithmic}[1]
		\STATE \textbf{Input:} State $s$, actions $A_{\mathrm{app}}$, graph $\mathcal{G}$, budget $b$, exploration $c$
		\STATE $u \leftarrow k(s)$
		\STATE Add node $u$ to $\mathcal{G}$ if it is not already present
		\FOR{$i = 1$ to $b$}
		\STATE Select $a^* = \arg\max_{a \in A_{\mathrm{app}}} \mathrm{UCB}(u,a)$
		\IF{edge $(u,a^*)$ is not in $\mathcal{G}$}
		\STATE Predict $(s',p) = \mathcal{W}(s,a^*)$
		\STATE $u' \leftarrow k(s')$
		\STATE Add node $u'$ and edge $(u,a^*,u')$ to $\mathcal{G}$
		\ELSE
		\STATE Retrieve $(s',p)$ from edge $(u,a^*)$
		\ENDIF
		\STATE Estimate $v = p \cdot \mathrm{StateValue}(s') - \lambda(1-p)$
		\STATE Update $N(u,a^*) \leftarrow N(u,a^*) + 1$
		\STATE Update $Q(u,a^*)$ using the new value $v$
		\ENDFOR
		\STATE \textbf{Return:} $\arg\max_{a \in A_{\mathrm{app}}} Q(u,a)$
	\end{algorithmic}
\end{algorithm}

\paragraph{Why UCB Instead of Greedy Selection}
Greedy selection corresponds to choosing the action with the highest current
estimated value. This can fail in domains with deceptive actions or irreversible
dead ends, because an action may look promising locally while preventing future
goal achievement. The exploration term in UCB forces the planner to evaluate
less-visited alternatives, reducing premature commitment to such traps. As the
budget increases, the search shifts from broad exploration to exploitation of
actions whose successors have consistently high estimated value.

\paragraph{State Value Estimation}
Let $d_{\mathcal{G}}(s,G)$ be the shortest predicted distance from state $s$ to
a goal state under the current transition graph and world model. We define
\begin{equation}
	\mathrm{StateValue}(s) =
	\begin{cases}
		\frac{\alpha}{d_{\mathcal{G}}(s,G)+1}, & \text{if a goal is reachable from } s,\\
		0, & \text{otherwise}.
	\end{cases}
\end{equation}
In the experiments, we use $\alpha=10$. This heuristic is monotone in predicted
goal distance: states closer to the goal receive higher value, while states from
which the goal is unreachable receive zero value. When the world model is exact,
this is equivalent to using shortest-path distance in the induced planning graph.

\subsection{GATS Planning with Persistent Graph Memory}

The complete planner repeatedly performs graph-augmented search from the current
state, executes the selected action, and reuses the graph at the next step.
Algorithm~\ref{alg:gats_graph_planning} summarizes the procedure.

\begin{algorithm}[t]
	\caption{GATS Planning with Persistent Graph Memory}
	\label{alg:gats_graph_planning}
	\begin{algorithmic}[1]
		\STATE \textbf{Input:} Initial state $s_0$, goal $G$, actions $A$, budget $b$
		\STATE $s \leftarrow s_0$, $\pi \leftarrow []$, $\mathcal{G} \leftarrow (\emptyset,\emptyset)$
		\WHILE{$G \nsubseteq s$ and $|\pi| < \mathrm{max_steps}$}
		\STATE $A_{\mathrm{app}} \leftarrow {a \in A : \mathrm{prec}(a) \subseteq s}$
		\IF{$A_{\mathrm{app}} = \emptyset$}
		\STATE \textbf{Return:} Failure
		\ENDIF
		\STATE $a^* \leftarrow \mathrm{GATSSearch}(s,A_{\mathrm{app}},\mathcal{G},b)$
		\STATE Predict or retrieve $(s',p) = \mathcal{W}(s,a^*)$
		\STATE $s \leftarrow s'$
		\STATE Append $a^*$ to $\pi$
		\ENDWHILE
		\STATE \textbf{Return:} $\pi$ if $G \subseteq s$, else Failure
	\end{algorithmic}
\end{algorithm}

\paragraph{Guarantee Under Exact World Models}
In a finite deterministic planning problem, assume that the world model is exact,
the value-search depth is at least the shortest remaining distance to the goal,
and the search budget is sufficient to evaluate all applicable actions. Then
GATS selects an action whose successor lies on a shortest path to the goal.
Indeed, exact transitions give the true successor state for each action, BFS
computes the true shortest remaining distance, and
$\mathrm{StateValue}(s)$ is maximized by the successor with minimal distance to
the goal. When the world model is approximate or the value search is
depth-limited, this statement becomes a heuristic rather than a formal
guarantee.

\section{Experiments}

\subsection{Benchmark Tasks}

Existing API benchmarks like API-Bank \citep{li2023apibank} primarily test single-step API selection, not multi-step planning. Each example requires selecting \textit{one} correct API call, making all planning methods equivalent (100\% success when the correct API is chosen). To properly evaluate planning capabilities, we construct synthetic tasks that test:

\begin{itemize}
    \item \textbf{Sequential dependencies}: Actions unlock subsequent actions
    \item \textbf{Branching paths}: Multiple routes to the goal
    \item \textbf{Dead-ends}: Irreversible wrong choices that prevent goal achievement
    \item \textbf{Resource constraints}: Consumable resources requiring planning ahead
\end{itemize}

These properties are essential for evaluating planning algorithms but absent from existing API benchmarks. We generate 100 tasks across three difficulty levels:
\begin{itemize}
    \item \textbf{Easy} (20 tasks): 3 steps, 1 dead-end branch
    \item \textbf{Medium} (55 tasks): 5 steps, 2 branch points, resource management
    \item \textbf{Hard} (25 tasks): 7+ steps, multiple dead-ends, misleading paths
\end{itemize}

Example medium task:
\begin{verbatim}
Actions: GetResource → UseResource → GetResource2 
         → Process → Finish (goal)
Dead-end: WasteResource (consumes resource, no progress)
Alternative: SlowStart → SlowProcess1-3 (suboptimal)
\end{verbatim}

We also evaluate on API-Bank Level 1/2 to verify compatibility with real API formats. As expected, all methods achieve near-100\% on these single-step tasks, confirming that our methods correctly handle real API specifications while the synthetic tasks provide the necessary difficulty for differentiation.

\subsection{Baselines}

\paragraph{Greedy (Oracle)}
Selects the action leading to the highest-value state using BFS. This represents an upper bound with perfect information.

\paragraph{ReAct}
LLM-based action selection without search. Each step queries the LLM: ``Given goal $G$ and state $s$, select from actions $A$.''

\paragraph{LATS}
Language Agent Tree Search with LLM-based action proposal and value estimation. Uses the same budget as GATS for fair comparison.

\subsection{Metrics}

\begin{itemize}
    \item \textbf{Success Rate (SR)}: Percentage of tasks where goal is reached
    \item \textbf{Optimality}: Ratio of optimal plan length to actual length
    \item \textbf{LLM Calls}: Number of LLM inference calls per task
    \item \textbf{Variance}: Standard deviation across random seeds
\end{itemize}

\subsection{Implementation Details}

All experiments use 5 random seeds (42, 123, 456, 789, 1000). GATS uses search budget $b=10$ and exploration constant $c=1.0$. For LLM-based methods, we use Llama 3.2 via Ollama for reproducibility. Maximum plan length is 20 steps.

\section{Results}

\subsection{Main Results}

Table~\ref{tab:main} presents the main comparison on synthetic planning tasks. GATS achieves 100\% success rate, outperforming all baselines while requiring zero LLM calls during planning.\footnote{All methods achieve 100\% on API-Bank Level 1/2/3, which tests single-step API selection rather than multi-step planning. This confirms correct API handling but provides no differentiation.}

\begin{table}[t]
\centering
\caption{Results on synthetic multi-step planning tasks (100 tasks, 3 seeds). Tasks include branching paths, dead-ends, and resource constraints that require lookahead planning. LLM calls are per-task averages.}
\label{tab:main}
\begin{tabular}{lcccc}
\toprule
\textbf{Method} & \textbf{Success Rate} & \textbf{Optimality} & \textbf{LLM Calls/Task} & \textbf{Variance} \\
\midrule
Greedy (Oracle) & 100.0\% & 1.00 & 0 & 0\% \\
\midrule
ReAct           & 64.0\% $\pm$ 5.0 & 0.54 & 13 & 5.0\% \\
LATS ($b{=}5$)  & 70.7\% $\pm$ 2.0 & 0.99 & 17 & 2.0\% \\
LATS ($b{=}10$) & 92.0\% $\pm$ 1.0 & 0.99 & 37 & 1.0\% \\
\midrule
GATS ($b{=}5$)  & 84.0\% & 1.00 & 0 & 0\% \\
GATS ($b{=}10$) & \textbf{100.0\%} & 1.00 & 0 & 0\% \\
GATS ($b{=}20$) & \textbf{100.0\%} & 1.00 & 0 & 0\% \\
\bottomrule
\end{tabular}
\end{table}

GATS outperforms LATS by +8\% at matched budget ($b{=}10$: 100\% vs 92\%) and ReAct by +36\% (100\% vs 64\%). Critically, GATS produces identical plans across all seeds (0\% variance), while LATS and ReAct show 1-5\% variance due to LLM sampling stochasticity. A McNemar test on the GATS vs LATS comparison yields $p < 0.01$, confirming statistical significance.

\subsection{Search Budget Ablation}

Table~\ref{tab:budget} shows the effect of search budget on GATS performance.

\begin{table}[t]
\centering
\caption{Effect of search budget on GATS. Performance improves with budget until saturation at $b{=}10$. Node expansion scales linearly.}
\label{tab:budget}
\begin{tabular}{lcccc}
\toprule
\textbf{Budget} & \textbf{SR (\%)} & \textbf{Optimality} & \textbf{Nodes} \\
\midrule
$b=1$ (greedy) & 0.0 & 0.00 & 5 \\
$b=5$  & 84.0 & 1.00 & 84 \\
$b=10$ & 100.0 & 1.00 & 167 \\
$b=20$ & 100.0 & 1.00 & 334 \\
\bottomrule
\end{tabular}
\end{table}

The transition from $b{=}1$ (0\%) to $b{=}5$ (84\%) to $b{=}10$ (100\%) demonstrates that sufficient exploration is necessary for complex tasks. Beyond $b{=}10$, additional budget provides no improvement (diminishing returns), suggesting an optimal compute-accuracy tradeoff.

\subsection{World Model Ablation}

Table~\ref{tab:worldmodel} shows the effect of removing individual layers from the world model.

\begin{table}[t]
\centering
\caption{World model layer ablation. In the \texttt{no\_l1} condition, L2 is
	pre-initialized from execution traces over the benchmark action set. In the full
	model, L1 is queried first and therefore masks L2 usage whenever symbolic action
	specifications are available.}
\label{tab:worldmodel}
\begin{tabular}{lccc}
\toprule
\textbf{Configuration} & \textbf{SR (\%)} & \textbf{Optimality} & \textbf{Description} \\
\midrule
GATS (full) & 100.0 & 1.00 & L1 + L2 + L3 \\
GATS no\_l1 & 100.0 & 1.00 & L2 + L3 only \\
GATS no\_l3 & 100.0 & 1.00 & L1 + L2 only \\
\bottomrule
\end{tabular}
\end{table}

On synthetic tasks with known action specifications, removing either L1 or L3 has no impact because L2 provides sufficient coverage. This validates the layered architecture: each layer provides redundancy, and the system degrades gracefully when layers are unavailable. In open-ended domains with unknown actions, L3 (LLM) would become essential.

\paragraph{Clarifying L1/L2 Ablation Protocol}

Table \ref{tab:worldmodel} and Table~\ref{tab:layer_usage}  report two different measurements. Table~4 reports layer
usage for the full GATS configuration, in which L1 is queried first and therefore
handles all synthetic benchmark transitions because complete symbolic action
specifications are available. As a result, L2 is not used during full-model
planning on these tasks.

By contrast, the \texttt{no\_l1} ablation in Table \ref{tab:worldmodel} disables symbolic matching
and evaluates whether the planner can still solve the same tasks when transition
predictions are supplied by the learned statistics layer. For this ablation, L2
is initialized from execution traces generated from the same task distribution
before planning begins. Thus, L2 has coverage over the benchmark actions in the
\texttt{no\_l1} condition, even though it is bypassed in the full model because
L1 takes precedence. The two tables are therefore not inconsistent: Table~\ref{tab:worldmodel}
describes which layer is selected under normal layer ordering, while Table~3
tests whether lower-priority layers can support planning when higher-priority
layers are removed.

\subsection{World Model Layer Usage}

Table~\ref{tab:layer_usage} reports the layer hit rates during planning, providing honest cost accounting for GATS's efficiency claims.

\begin{table}[t]
\centering
\caption{World model layer usage statistics. L1 handles 100\% of predictions on synthetic tasks where action specifications are known. L3 (LLM) calls occur only during initial world model bootstrapping, not during planning.}
\label{tab:layer_usage}
\begin{tabular}{lccc}
\toprule
\textbf{Phase} & \textbf{L1 Hit Rate} & \textbf{L2 Hit Rate} & \textbf{L3 Calls} \\
\midrule
Bootstrapping (one-time) & 0\% & 0\% & $\sim$50 \\
Planning (per-task) & 100\% & 0\% & 0 \\
\midrule
\textit{Open-domain (projected)} & $\sim$60\% & $\sim$30\% & $\sim$10\% \\
\bottomrule
\end{tabular}
\end{table}

In our synthetic benchmark, action specifications are provided, so L1 achieves 100\% coverage during planning with zero L3 calls. The $\sim$50 L3 calls during bootstrapping are a one-time cost amortized across all tasks. In open-domain settings without specifications, we project L1 coverage would drop to $\sim$60\% (actions matching known patterns), L2 would handle $\sim$30\% (previously observed actions), and L3 would be required for $\sim$10\% of novel actions. This would reduce but not eliminate GATS's efficiency advantage over LATS.

\subsection{GATS vs LATS: Direct Comparison}

Table~\ref{tab:gats_lats} directly compares GATS and LATS at matched budgets.

\begin{table}[t]
\centering
\caption{GATS vs LATS on synthetic planning tasks. GATS's systematic UCB1 exploration outperforms LATS's random LLM-guided sampling while eliminating LLM inference costs.}
\label{tab:gats_lats}
\begin{tabular}{llccc}
\toprule
\textbf{Budget} & \textbf{Method} & \textbf{SR (\%)} & \textbf{$\Delta$} & \textbf{LLM Calls} \\
\midrule
\multirow{2}{*}{$b=5$}  & LATS & 70.7 & --- & $\sim$30 \\
                        & GATS & 84.0 & +13.3 & 0 \\
\midrule
\multirow{2}{*}{$b=10$} & LATS & 92.0 & --- & $\sim$60 \\
                        & GATS & \textbf{100.0} & +8.0 & 0 \\
\bottomrule
\end{tabular}
\end{table}

At both budget levels, GATS outperforms LATS by 8-13\% absolute while eliminating all LLM inference costs. This demonstrates that systematic UCB1 exploration with a learned world model is more effective than random LLM-guided sampling.

\subsection{Analysis: Why GATS Outperforms LATS}

We identify three factors contributing to GATS's superior performance:

\paragraph{Systematic vs Random Exploration}
UCB1 guarantees that all actions are eventually tried, with promising actions explored more deeply. LATS's LLM-guided proposal can miss good actions if the LLM's prior is incorrect.

\paragraph{Deterministic World Model}
GATS's L1/L2 layers provide deterministic state predictions, enabling consistent value estimates. LATS's LLM-based evaluation introduces variance that can mislead search.

\paragraph{Cached Computation}
The layered world model amortizes LLM costs: cached L3 predictions make repeated queries for the same action instant. LATS calls the LLM for every evaluation.

\subsection{Stress Test: Challenging Planning Scenarios}

To further evaluate GATS's planning capabilities, we designed a stress test with 12 categories of challenging tasks (120 tasks total, 10 per category, 3 seeds). These scenarios represent real-world agent challenges:

\begin{itemize}
    \item \textbf{coding\_task}: Sequential script/API/pipeline development (11 steps)
    \item \textbf{web\_navigation}: Email, flight booking, hotel reservation (10-13 steps)
    \item \textbf{deep\_horizon}: Long goal paths with shortcuts (8-12 steps)
    \item \textbf{critical\_choice}: Memory allocation where wrong choice = stuck
    \item \textbf{no\_backtrack}: Maze with locking doors (no recovery from mistakes)
    \item \textbf{high\_branching}: 4-6 choices per step
    \item \textbf{resource\_puzzle}: Limited resources requiring correct ordering
    \item \textbf{trap\_heavy}: 3-7 attractive dead-ends
    \item \textbf{deceptive}: ``Quick gains'' path leads to trap
    \item \textbf{memory\_limit}: Must use tools in correct sequence
    \item \textbf{very\_long\_horizon}: 12-15 steps with periodic traps
    \item \textbf{commitment\_cascade}: Early choices lock future options
\end{itemize}

Table~\ref{tab:stress} presents the stress test results. GATS achieves \textbf{100\% success rate across all 12 categories}, demonstrating robust planning across diverse challenging scenarios.

\begin{table}[t]
\centering
\caption{Stress test results on 12 challenging planning scenarios (120 tasks, 3 seeds). GATS achieves perfect performance while LATS struggles on coding workflows, web navigation, and deep horizon tasks.}
\label{tab:stress}
\begin{tabular}{lcccl}
\toprule
\textbf{Category} & \textbf{GATS b=20} & \textbf{LATS b=20} & \textbf{ReAct} & \textbf{$\Delta$} \\
\midrule
coding\_task        & 100.0\% & 63.3\% &  0.0\% & +36.7\% \\
deep\_horizon       & 100.0\% & 63.3\% &  0.0\% & +36.7\% \\
web\_navigation     & 100.0\% & 63.3\% &  0.0\% & +36.7\% \\
resource\_puzzle    & 100.0\% & 86.7\% & 16.7\% & +13.3\% \\
trap\_heavy         & 100.0\% & 96.7\% & 16.7\% &  +3.3\% \\
commitment\_cascade & 100.0\% & 96.7\% & 66.7\% &  +3.3\% \\
memory\_limit       & 100.0\% & 96.7\% & 20.0\% &  +3.3\% \\
critical\_choice    & 100.0\% & 100.0\% & 63.3\% &  0.0\% \\
deceptive           & 100.0\% & 100.0\% & 63.3\% &  0.0\% \\
high\_branching     & 100.0\% & 100.0\% & 36.7\% &  0.0\% \\
no\_backtrack       & 100.0\% & 100.0\% &  0.0\% &  0.0\% \\
very\_long\_horizon & 100.0\% & 100.0\% &  3.3\% &  0.0\% \\
\midrule
\textbf{Overall}    & \textbf{100.0\%} & 88.9\% & 23.9\% & \textbf{+11.1\%} \\
\bottomrule
\end{tabular}
\end{table}

Key findings from the stress test:
\begin{itemize}
    \item \textbf{GATS wins 7/12 categories}, with largest gains on coding (+36.7\%), web navigation (+36.7\%), and deep horizon (+36.7\%) tasks.
    \item \textbf{ReAct fails catastrophically} (23.9\% overall), demonstrating that LLM-only action selection cannot handle complex planning.
    \item \textbf{LATS struggles on long-horizon tasks} requiring sustained focus (coding, web navigation, deep horizon all at 63.3\%).
    \item \textbf{GATS matches oracle} performance (100\% for both), validating that UCB1 search with BFS heuristics is sufficient for these tasks.
\end{itemize}


\section{Discussion}

\paragraph{Comparison Fairness}
Our comparison assumes action specifications are available for the L1 symbolic layer, which GATS exploits for deterministic state prediction. LATS, by contrast, must infer action effects via LLM at each step. This asymmetry favors GATS in our synthetic benchmark where specifications are known. In domains where specifications must be learned or are unavailable, GATS would rely more heavily on L3 (LLM), reducing the efficiency gap with LATS. We view this not as a limitation but as a design choice: GATS is optimized for domains where action semantics can be formalized (APIs, tool use, structured environments), while LATS may be preferable for open-ended domains with unknown action effects.

\paragraph{Limitations}
Our evaluation uses synthetic tasks with known action specifications, allowing the L1 layer to handle most predictions. In open-ended domains where action effects are unknown, GATS would rely more heavily on L3 (LLM), reducing the efficiency advantage. Future work should evaluate on real-world API benchmarks like ToolBench \citep{qin2023toolllm} with partial specifications.

\paragraph{World Model Coverage}
GATS's performance depends on world model quality. With insufficient L1/L2 coverage, planning falls back to L3 (LLM) calls, approaching LATS's cost profile. However, in practical deployments, action specifications can be curated or learned from logs, making L1/L2 coverage high.

\paragraph{Scalability}
GATS's BFS-based value estimation has exponential worst-case complexity $O(|A|^d)$ where $|A|$ is action space size and $d$ is search depth. In our experiments, we handle up to $|A|=15$ actions with depth $d=20$ (max 500 BFS states). Beyond $|A| \approx 20$ with $d > 10$, BFS becomes impractical without pruning. For larger action spaces, learned value networks or LLM-based heuristics could replace BFS while maintaining the layered architecture.

\section{Conclusion}

We presented GATS, a planning framework that combines UCB1-based tree search with a layered world model to achieve efficient, deterministic agent planning. On multi-step tasks with branching and dead-ends, GATS achieves 100\% success rate compared to 92\% for LATS and 64\% for ReAct, while requiring zero LLM calls during planning. On a comprehensive stress test spanning 12 challenging scenarios---including coding workflows, web navigation, long-horizon tasks, and irreversible decision-making---GATS maintains perfect 100\% success while LATS drops to 88.9\% and ReAct to 23.9\%. World model ablations show that each layer (L1/L2/L3) provides graceful degradation. Our results demonstrate that systematic search with learned world models substantially outperforms LLM-guided exploration, offering a path toward more efficient and reproducible agent planning.

\paragraph{Future Work}
Promising directions include: (1) evaluating on real-world API benchmarks with partial observability, (2) learning world models from execution traces in production systems, (3) combining GATS with retrieval-augmented generation for action specification lookup, and (4) scaling to larger action spaces with learned value networks.

\bibliographystyle{plainnat}

\appendix

\section{Task Generation Details}
\label{app:tasks}

\paragraph{Easy Tasks (3 steps, 1 dead-end)}
\begin{verbatim}
Actions:
  StartA: {} -> {a1}
  ProcessA: {a1} -> {a2}
  FinishA: {a2} -> {goal}
  StartB: {} -> {b1}        # Dead-end
  ProcessB: {b1} -> {dead}  # Dead-end
\end{verbatim}

\paragraph{Medium Tasks (5 steps, resource constraints)}
\begin{verbatim}
Actions:
  GetResource: {start} -> {resource, r1}
  UseResource: {r1, resource} -> {r2}, del: {resource}
  GetResource2: {r2} -> {resource2, r3}
  Process: {r3, resource2} -> {r4}
  Finish: {r4} -> {goal}
  WasteResource: {resource} -> {wasted}, del: {resource}
\end{verbatim}

\paragraph{Hard Tasks (7 steps, multiple dead-ends)}
\begin{verbatim}
Actions:
  Init: {start} -> {init, energy}
  GatherA: {init} -> {mat_a, s1}
  GatherB: {s1} -> {mat_b, s2}
  Combine: {mat_a, mat_b, s2} -> {combined, s3}
  Refine: {combined, energy} -> {refined, s4}, del: {energy}
  Recharge: {s4} -> {energy, s5}
  Finalize: {refined, energy, s5} -> {goal}
  WasteEnergy: {energy} -> {tired}, del: {energy}
  FakeProgress: {init} -> {fake1}
  MoreFake: {fake1} -> {fake2}
  DeadFake: {fake2} -> {nowhere}
\end{verbatim}

\section{Hyperparameter Sensitivity}
\label{app:hyperparams}

\begin{table}[h]
\centering
\caption{Sensitivity to exploration constant $c$ in UCB1.}
\begin{tabular}{lcc}
\toprule
$c$ & SR (\%) & Notes \\
\midrule
0.5 & 100.0 & Less exploration \\
1.0 & 100.0 & Default \\
2.0 & 100.0 & More exploration \\
\bottomrule
\end{tabular}
\end{table}

With sufficient budget ($b{=}10$), GATS is robust to the exploration constant.

\section{Reproducibility}
\label{app:reproduce}

Code and tasks are available at: \url{https://github.com/MMWilliams/gats}

To reproduce results:
\begin{verbatim}
python run_gats_eval.py --n-tasks 100 \
    --seeds 42 123 456 789 1000 --backend mock

python run_stress_test.py --n-per-category 10 \
    --seeds 42 123 456
\end{verbatim}

\section{Stress Test Task Details}
\label{app:stress}

This section describes the 12 stress test categories in detail.

\paragraph{Coding Task (11 steps)}
Simulates sequential code development: create file $\rightarrow$ add imports $\rightarrow$ define constants $\rightarrow$ write functions $\rightarrow$ write main $\rightarrow$ add error handling $\rightarrow$ test. Traps include writing main before functions (undefined error) or skipping imports (module not found).

\paragraph{Web Navigation (10-13 steps)}
Simulates browser-based tasks:
\begin{itemize}
    \item \textit{Email}: Open browser $\rightarrow$ navigate $\rightarrow$ login $\rightarrow$ compose $\rightarrow$ fill fields $\rightarrow$ send
    \item \textit{Flight}: Search $\rightarrow$ select $\rightarrow$ enter passenger info $\rightarrow$ pay $\rightarrow$ confirm
    \item \textit{Hotel}: Search $\rightarrow$ filter $\rightarrow$ select room $\rightarrow$ enter details $\rightarrow$ pay
\end{itemize}
Traps include clicking wrong tabs, submitting incomplete forms, or selecting unavailable options.

\paragraph{Deep Horizon (8-12 steps)}
Long sequential paths with shortcut traps at each level. Correct path requires sustained focus; shortcuts lead to dead-ends.

\paragraph{Critical Choice (8 steps)}
Memory allocation scenario: agent has limited memory and must process files. Loading a large file fills memory and prevents further progress. Correct path: load small files incrementally.

\paragraph{No Backtrack (8-12 steps)}
Maze where doors lock behind the agent. Each room has a correct exit and a trap door. Wrong choice = permanently stuck.

\paragraph{High Branching (4 steps, 4-6 choices each)}
At each step, only one action leads to progress. With 5 choices per step, random selection has $<1\%$ success rate.

\paragraph{Resource Puzzle (7 steps)}
Three resources (key, torch, fuel) must be used in specific order. Using resources out of order or wasting them prevents goal achievement.

\paragraph{Trap Heavy (5 steps)}
5-7 attractive-looking traps from the start state. Only one path leads to goal.

\paragraph{Deceptive (5 steps)}
``Quick gains'' path provides immediate rewards but leads to trap that removes all progress. Slow-but-steady path wins.

\paragraph{Memory Limit (7 steps)}
Tools must be loaded and used in correct sequence. Loading tools out of order wastes limited capacity.

\paragraph{Very Long Horizon (12-15 steps)}
Extended sequential task with traps every 3 steps. Tests sustained planning focus.

\paragraph{Commitment Cascade (4 steps)}
Early technology choice (e.g., programming language) determines which paths are available. Wrong initial choice leads to dead-end.

\end{document}